\documentclass[conference]{IEEEtran}
\IEEEoverridecommandlockouts

\usepackage{booktabs}
\usepackage{cite}
\usepackage{amsmath,amssymb,amsfonts}
\usepackage{algorithmic}
\usepackage{graphicx}
\usepackage{textcomp}
\usepackage{xcolor}
\def\BibTeX{{\rm B\kern-.05em{\sc i\kern-.025em b}\kern-.08em
    T\kern-.1667em\lower.7ex\hbox{E}\kern-.125emX}}

\begin{document}

\title{\LARGE \bf
Vision--Based Leader--Follower Formation Control for Cooperative UAVs in GPS--Degraded Environments
}

\author{
\IEEEauthorblockN{
Deekshitha Angadi\textsuperscript{1,2},
Naveena Budda\textsuperscript{3},\\
Vikas Agarwal\textsuperscript{4,5},
Rojesh Arunkumar Mulasa\textsuperscript{6},
Ravi Killamsetty\textsuperscript{1},\\
Mohamed Samshad\textsuperscript{7},
Narsimlu Kemsaram\textsuperscript{8}
}
\IEEEauthorblockA{
\textsuperscript{1}\textit{AI, IoT and Robotics Lab (AIR Lab), UAVs Group, Autonomous Robotics Systems Limited, Hyderabad, India}\\
\textsuperscript{2}\textit{Department of Microelectronics and VLSI Design, University of Hyderabad, Hyderabad, India}\\
\textsuperscript{3}\textit{Department of Internet of Things, Ideabytes Software India Private Limited, Hyderabad, India}\\
\textsuperscript{4}\textit{School of Computer Science, Georgia Institute of Technology, Atlanta, USA}\\
\textsuperscript{5}\textit{Enterprise Solutions Unit, Tata Consultancy Services, Atlanta, USA}\\
\textsuperscript{6}\textit{Mobile Application Development, Tata Consultancy Services, Tokyo, Japan}\\
\textsuperscript{7}\textit{Department of Electrical Engineering, Indian Institute of Technology, Kanpur, India}\\
\textsuperscript{8}\textit{Department of Artificial Intelligence, University of Malaya, Kuala Lumpur, Malaysia}
}
}

\maketitle
\thispagestyle{empty}
\pagestyle{empty}

\begin{abstract}


Cooperation in multi-UAV systems requires reliable relative perception so that follower vehicles can maintain formation and continue their mission safely even when absolute-positioning sensors degrade or fail. 
This paper presents a vision-based cooperative formation framework running on a follower UAV that uses a front-facing RGB-D camera to detect, track, and localize a leader UAV in real-time. A lightweight YOLO-based detector is trained on a dedicated drone dataset and deployed onboard to predict leader bounding boxes, which are then fused with depth information via a pinhole camera model to estimate the leader's relative pose. These estimates provide a leader-follower position controller and can also be used as a backup when GPS or external localization is unavailable. This framework is implemented as a set of ROS nodes and evaluated in a physics-based multi-UAV simulation built on XTDrone, with sensor noise and communication dropouts. 
We evaluate detection accuracy, runtime, and formation-keeping error under nominal conditions and under simulated failures of the positioning sensors. 
The results show that the proposed framework maintains stable leader-follower formations with reasonable computational cost and provides a practical basis for extending vision-based cooperative formation control to real-world multi-UAV systems.

\end{abstract}


\section{ Introduction }


Unmanned aerial vehicles (UAVs) are increasingly used for cooperative tasks such as inspection \cite{saha2023autonomous}, surveillance \cite{kemsaram2014experimental}, precision agriculture \cite{potena2019agricolmap}, search-and-rescue \cite{queralta2020collaborative}, disaster response \cite{gregory2016application}, defense emergence \cite{deka2021natural}, and space exploration \cite{das2023stixel}. 
In many of these scenarios \cite{cao2025cooperative}, leader-follower formations provide a simple and effective means of imposing structure on the group: one vehicle follows a reference trajectory, and the others maintain prescribed relative positions and headings.
Traditional formation control strategies often rely heavily on absolute positioning systems, such as the global positioning system (GPS) \cite {li2014localization} or motion capture (MoCap) system \cite{rafifandi2019leader}, to determine each vehicle's pose. 
In cluttered or GPS-denied environments, however, such assumptions are fragile \cite{ait2025distributed}. 
To enable robust cooperation, follower UAVs require local onboard perception that allows them to estimate the relative pose of their neighbours using only their own sensors.

\begin{figure}[!t]
    \centering
    \includegraphics[width=0.45\textwidth]{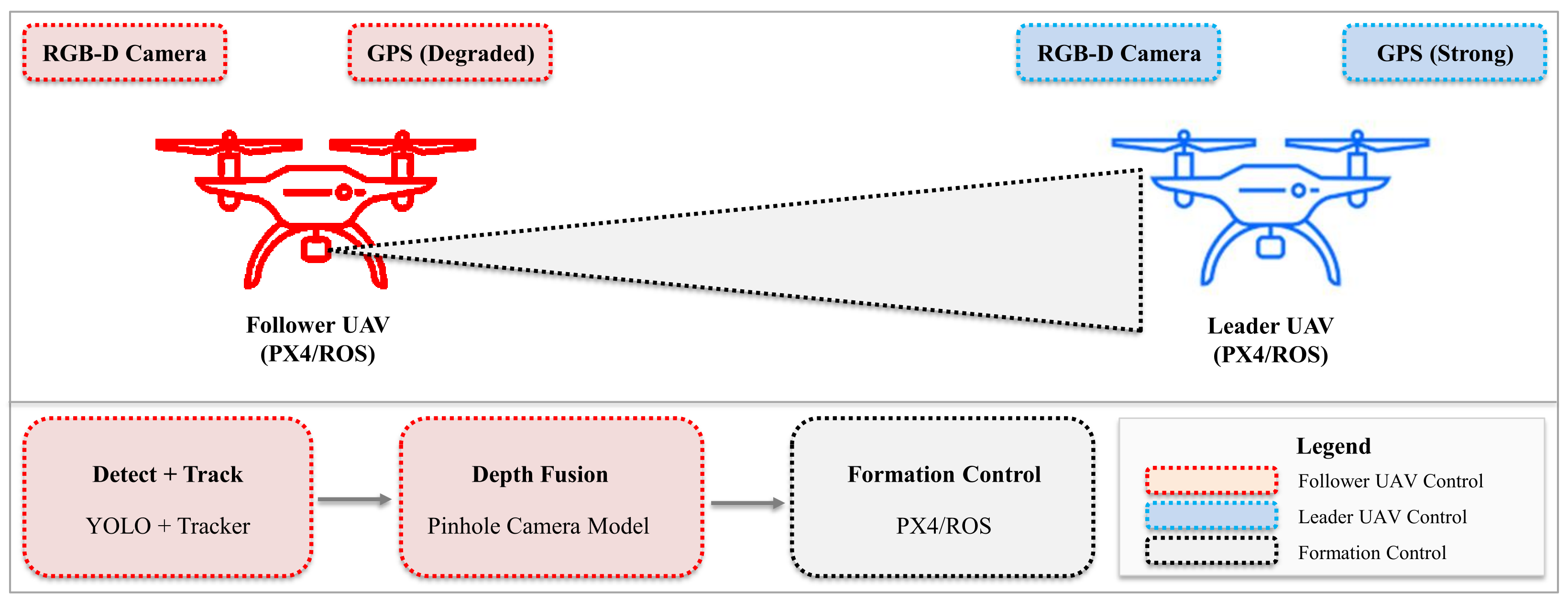}
    \caption{ 
    Vision-based leader–follower formation control with degraded positioning on the follower UAV. 
    }
    \label{Figure:CUAV_BlockDiagram}
\end{figure}

In this paper, we explore a vision-based approach to leader-follower formations, focusing on a minimal but representative setting with one leader and one follower. 
Figure \ref{Figure:CUAV_BlockDiagram} shows the block diagram of the vision-based leader-follower formation framework for cooperative UAVs.
The follower is equipped with an RGB-D camera and a lightweight perception pipeline that detects the leader in the image stream and reconstructs its relative pose using depth information and a calibrated pinhole camera model. This relative pose estimate is then used to maintain formation and to provide a safety layer in the event of failures in the positioning sensors.

The main contributions of this paper are: i) a modular, vision-based perception framework for leader--follower formation, based on a custom-trained YOLOv8 drone detector and RGB-D geometry, ii) a complete implementation in ROS integrated with XTDrone and PX4 SITL on Gazebo, suitable for multi-UAV simulation, and iii) an evaluation of detection accuracy, pose-estimation error, and formation-keeping performance under nominal and sensor-failure conditions.





\section{ Related Work } \label{Sec_RelatedWork}


This section presents a concise review of leader–follower formation, vision-based relative localization, and deep-learning-based drone detection algorithms. 

\subsection{Cooperative UAVs and Leader-Follower Formation}

Surveys on cooperative UAVs summarize key architectures, coordination strategies, and enabling technologies for multi-UAV systems in both civilian and defence applications \cite{kemsaram2016autonomous}, with leader–follower formation emerging as a widely adopted and flexible coordination pattern for missions \cite{olfati2006flocking}. 
Classical formation-control approaches, largely adapted from ground robotics, include virtual-structure, behaviour-based, and graph-theoretic methods \cite{ren2007information}. 
These methods can achieve stable formations under connectivity and collision-avoidance constraints \cite{kemsaram2017design}, but they become fragile in cluttered or GPS-degraded environments where absolute positioning is unreliable or unavailable \cite{guo2019ultra}. 

\subsection{Vision-Based Relative Localization and Formation}

Vision-based relative localization is well suited to UAVs because cameras are lightweight, low-cost, and provide rich scene and inter-agent information \cite{kemsaram2020architecture}.
Early leader–follower systems commonly used fiducial markers (e.g., AprilTag/ArUco) to obtain accurate short-range 6-DoF relative pose \cite{evangeliou2022visual}, enabling reliable indoor formation demonstrations, but requiring clear line-of-sight and environment information that does not scale well to outdoor missions \cite{olson2011apriltag}.
Markerless approaches instead estimate relative motion from natural imagery using feature matching with epipolar geometry \cite{rozsa2022immediate} or recover depth via stereo visual odometry \cite{forster2014svo}, and these estimates are typically fed into classical formation controllers to maintain relative spacing when GPS is degraded.

\subsection{Deep-Learning-Based Drone Detection}

The emergence of real-time object detectors has significantly improved the robustness of UAV detection in cluttered environments \cite{kemsaram2017multi}. 
Models based on the YOLO \cite{sahin2021yolodrone}, Faster R-CNN \cite{nalamati2019drone}, SSD \cite{topalli2021real} architectures have been trained specifically for drone detection in surveillance and sense-and-avoid applications, often using synthetic or augmented datasets to cover a wide range of scales, backgrounds, and viewing angles \cite{alkentar2021practical}. 
These detectors can recognise UAVs without fiducial markers and have been deployed on embedded platforms such as Nvidia Jetson, Raspberry Pi, Intel NUC boards. 


Our contribution is to bridge this gap by integrating a deep-learning-based drone detector, RGB-D sensing, and a leader-follower formation controller into a single, modular framework implemented in ROS and evaluated in the XTDrone-based (on Gazebo) simulation environment. 

\section{Proposed Vision-Based Leader-Follower Formation for Cooperative UAVs} \label{Sec_ProposedSystem}




We consider a minimal but representative cooperative setting with two multirotor UAVs: a leader, which follows a predefined mission trajectory using its onboard navigation stack, and a follower, which must maintain a desired formation offset pose relative to the leader. 
In conventional approaches, this relative pose is inferred primarily from GPS or MoCap systems. 
In contrast, we investigate a vision-based strategy in which the follower relies on its onboard camera and computation to estimate the leader's pose and respond robustly when absolute positioning information is degraded or lost.
The follower is equipped with a front-facing RGB-D camera and runs a cooperative perception pipeline that: i) acquires synchronized RGB and depth images, ii) detects the leader in the RGB frame using a custom-trained YOLOv8 drone model, iii) estimates the leader's relative pose from depth via a pinhole camera model, iv) filters the pose estimates over time, and v) forwards the relative pose to a leader-follower controller.


This architecture enables the follower to maintain formation when its positioning sensors are healthy and to fall back to vision-only relative pose estimation when they fail, either continuing the mission or executing a safe landing.

\subsection{Training Dataset}

To obtain a drone detector tuned to identify and track the leader UAV, we collected a drone dataset that combines publicly available UAV images with rendered views of PX4 drone. 
The dataset contains approximately 1,339 images spanning different altitudes, viewing angles, and distances, including both static and flying configurations. 
Each image was annotated with a tight bounding box around the leader drone using a Roboflow annotation tool.
We divided the dataset into a training split (80\%) and a testing split (20\%), ensuring that short temporal sequences were not split across sets. 



\subsection{Drone Detection}

Drone detection is performed by a single-shot detector based on a compact YOLOv8 architecture, chosen for its balance between detection accuracy and runtime on embedded hardware. 
The network processes an input RGB image and outputs bounding boxes, objectness scores, and class probabilities over a regular grid. 
Non-maximum suppression is applied to remove overlapping boxes. 
We then retain the highest-confidence bounding box associated with the ``drone'' class, provided that its confidence exceeds a fixed threshold.
\subsection{Pose Estimation}

\begin{figure}[!t]
    \centering
    \includegraphics[width=0.45\textwidth]{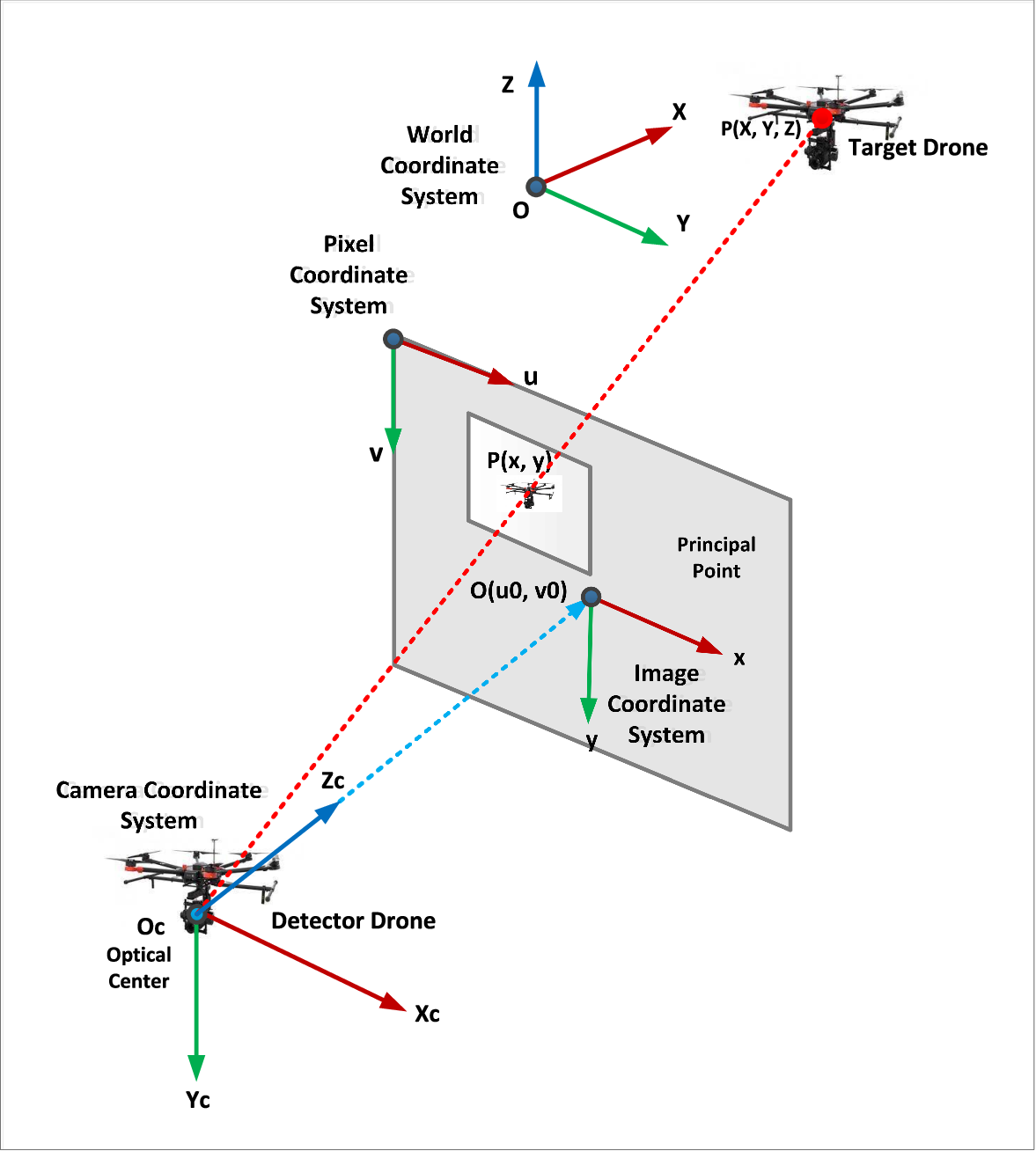}
    \caption{
    Pinhole camera geometry for vision-based relative pose estimation between UAVs.
    }
    \label{fig:CUAV_PoseEstimation}
\end{figure}

The follower's RGB-D camera supplies an RGB image and an aligned depth map for each frame. 
Let $(u, v)$ denote the pixel coordinates of the centre of the detected bounding box, and let $Z$ denote the depth at this location, computed as the median in a small patch around $(u, v)$ to mitigate noise. 
With the camera intrinsic matrix $K$, the corresponding 3D point in the camera frame is obtained via the pinhole camera model \cite{hartley2003multiple}:
\begin{equation}
    \begin{bmatrix} X \\ Y \\ Z \end{bmatrix}
    = K^{-1}
    \begin{bmatrix} u \\ v \\ 1 \end{bmatrix} Z
\end{equation}
This yields the leader's relative position $P(X,Y,Z)$ in the camera coordinate system.
The relative yaw between the leader and follower can be estimated from the follower's onboard heading, combined with the known extrinsic transform between the camera and vehicle frames. 
Figure \ref{fig:CUAV_PoseEstimation} illustrates the camera geometric model used for pose estimation, where $(X_c, Y_c, Z_c)$ represent the camera coordinate frame, and the image plane axes are aligned with the pixel coordinates.

\subsection{Temporal Filtering and Fusion}

The raw relative pose estimates derived from the pinhole camera model are affected by depth noise and occasional missed detections. 
To obtain a smoother, more reliable control signal, we apply a constant-velocity Kalman filter to the relative position. 
The filter predicts the state forward in time and corrects it upon receipt of a new vision-based measurement. 
When GPS or external position estimates are available, the vision-based relative pose fuses with these signals in a higher-level estimator. 
In this work, we focus on the fallback behaviour: when the follower's positioning sensor fails, the formation controller relies primarily on the estimated vision-based relative pose.

\subsection{Leader-Follower Formation}

The vision pipeline described above provides, at each time step, an estimate of the leader position in the follower's camera frame, 
\begin{equation}
    P(t) = (X(t), Y(t), Z(t))
\end{equation}
where, $Z$ is the distance along the optical axis and $X$, $Y$ are the lateral and vertical offsets. 
We define a desired formation configuration by specifying a target offset,
\begin{equation}
    P^{d}(t) = (X^{d}(t), Y^{d}(t), Z^{d}(t))
\end{equation}
which corresponds to the leader's position relative to the follower (e.g., centred in the image at a fixed distance in front of the camera).

At the control level, we work with a simple position error,
\begin{equation}
    e(t) = P(t) - P^{d}(t) = 
    \begin{bmatrix}
        e_X(t) \\ e_Y(t) \\ e_Z(t)
    \end{bmatrix}
    =
    \begin{bmatrix}
        X(t) - X^{d}(t) \\
        Y(t) - Y^{d}(t) \\
        Z(t) - Z^{d}(t)
    \end{bmatrix}
\end{equation}
and use this to generate a velocity command for the follower. 

We adopt a decoupled PID law of the form,
\begin{equation}
    v(t) =
    \begin{bmatrix}
        v_X(t) \\ v_Y(t) \\ v_Z(t)
    \end{bmatrix}
    =
    K_P\, e(t)
    + K_I \int_{0}^{t} e(\tau)\, d\tau
    + K_D\, \dot{e}(t)
\end{equation}
where, $K_P$, $K_I$, and $K_D$ are diagonal gain matrices.

In practice, we use proportional-derivative action in the horizontal plane with a small integral term to remove steady-state offsets, and proportional-integral control in altitude. 
The derivative term $\dot{e}(t)$ is obtained from finite differences of the filtered pose estimates, and the integral term is bounded to avoid windup when the commanded velocities saturate.
The resulting velocity command $v(t)$ is expressed in the local reference frame used by PX4 (e.g., a local NED) and sent as an offboard velocity setpoint via the XTDrone interface. 
PX4 then handles the inner-loop attitude and thrust control required to track these setpoints. 
When GPS or external localization is available, the autopilot maintains the local frame, and the formation controller regulates the relative offset to the leader.
In the event of a positioning sensor failure on the follower, the same PID law continues to operate using only the vision-based relative pose, allowing the follower either to maintain an approximate leader-follower formation or to decelerate and descend in a controlled manner.

This design keeps the high-level formation controller deliberately simple while remaining expressive enough to compensate for steady-state biases in the vision-based relative pose, thereby keeping the focus of the study on perception and its integration with an off-the-shelf flight stack.


\section{ Implementation } \label{Sec_SystemDevelopment}

\begin{figure}[!t]
    \centering
    \includegraphics[width=0.45\textwidth]{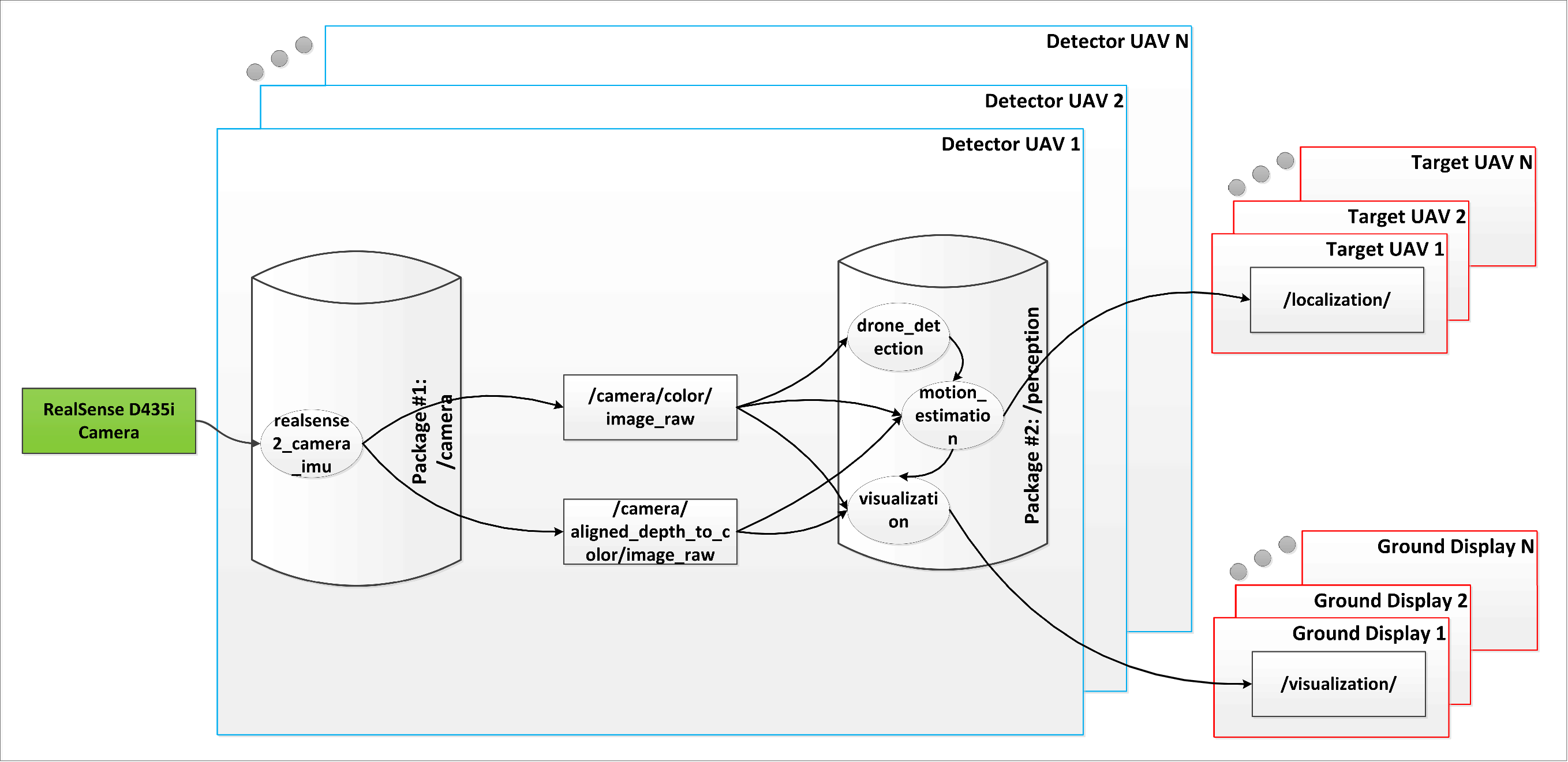}
    \caption{
    ROS node–topic architecture of the proposed vision-based leader–follower pipeline:
    The follower UAV subscribes to RGB-D streams from the RealSense D435i, performs leader detection and relative pose estimation, and publishes the estimated relative pose to the leader–follower position controller via MAVROS for closed-loop formation keeping.
    } \label{fig:CUAV_FrameworkDevelopment}
\end{figure}

The proposed perception framework is implemented in Python 3.8 using OpenCV 4.1 and PyTorch 1.9.0, and integrated with ROS Melodic on Ubuntu 18.04 LTS. 
Each logical component runs as a ROS node to simplify substitution and experimentation.
The RGB-D camera node publishes color and depth images aligned with the camera intrinsic parameters. 
The detection node subscribes to RGB images, executes the YOLOv8-based trained model, and publishes detected bounding boxes and confidence scores. 
A pose-estimation node subscribes to both the bounding boxes and depth images, applies the pinhole model and Kalman filter, and publishes relative pose estimates at approximately 20-25 Hz.
A controller bridge node converts relative pose estimates into velocity or position setpoints for the follower UAV and communicates these to the PX4 flight stack using MAVROS. Figure \ref{fig:CUAV_FrameworkDevelopment} shows the nodes and topics in the proposed system.
On a laptop with an Intel Core i7 CPU and an Nvidia GeForce RTX 3060 GPU, the end-to-end perception pipeline, including image acquisition, detection, and pose estimation, achieves an average processing time of 38 milliseconds per frame. 

%
\section{ Evaluation } \label{Sec_Evaluation}





\begin{figure}[!htbp]
    \centering
    \includegraphics[width=0.45\textwidth]{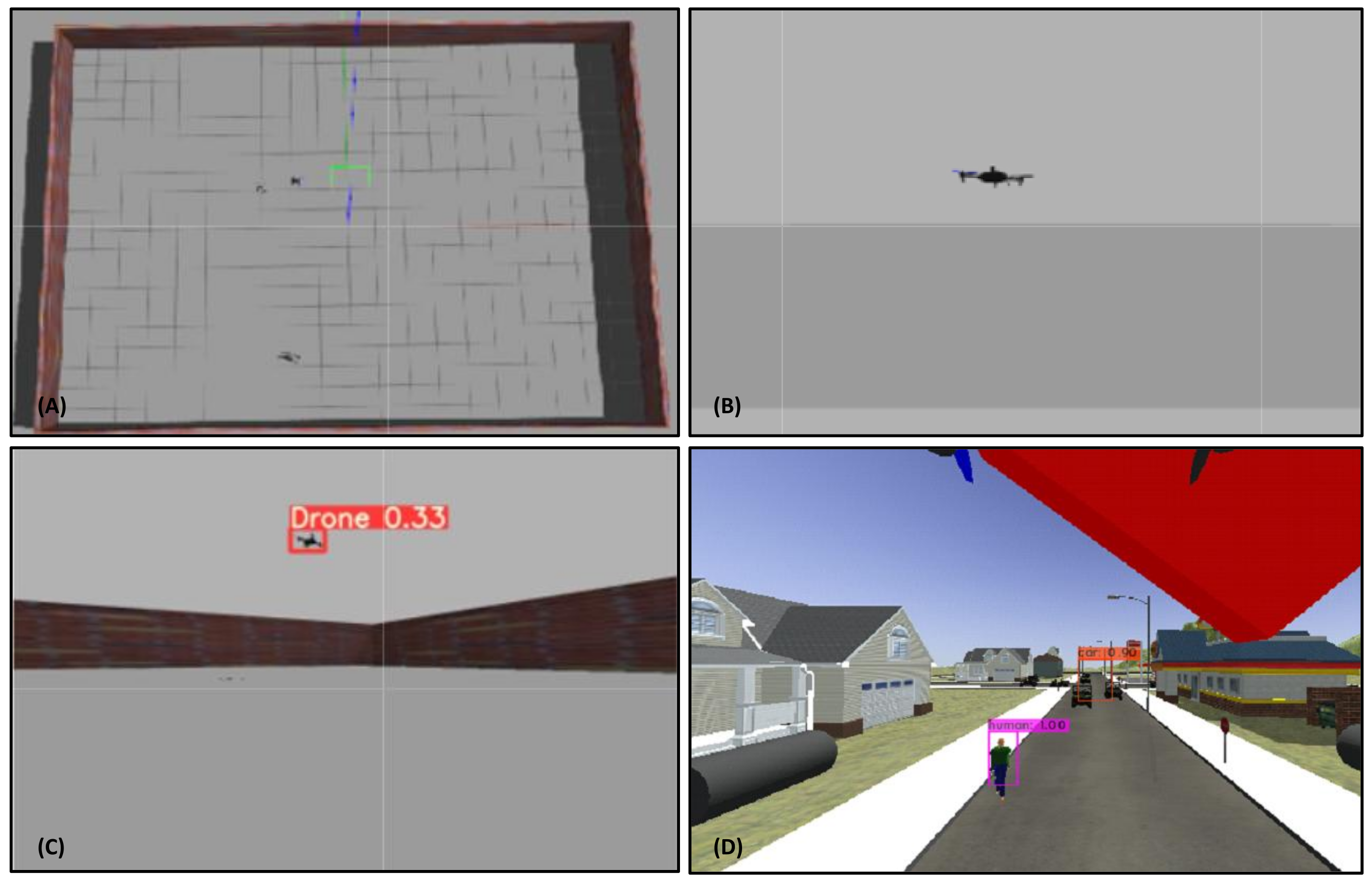}
    \caption{
    Multi-UAV simulation scenes in XTDrone/Gazebo used for evaluation: 
    (A) Leader-follower view of the two UAVs in the environment, 
    (B) Follower UAV camera view with the leader in the field of view,
    (C) Onboard detection result showing the leader UAV bounding box, and 
    (D) Outdoor flight scenario illustrating leader–follower tracking using the proposed RGB-D perception and relative pose estimation pipeline.
    }
    \label{fig:CUAV_SimulationResuts}
\end{figure}

To evaluate the proposed framework, we use a multi-UAV simulation built on XTDrone that integrates PX4 SITL, ROS, and Gazebo. 
The simulated world represents an open outdoor environment with textured ground and various structures to provide visual features (see Figure \ref{fig:CUAV_SimulationResuts}).
Two quadrotor UAVs are instantiated: a leader and a follower. 
The leader tracks predefined trajectories such as straight lines and circular paths at moderate speeds and altitudes. 
The follower is spawned nearby and switches to leader-follower mode once the perception system has detected the leader.
A virtual RGB-D camera, configured with the same field of view and resolution as a typical depth camera (Intel RealSense D435i), is rigidly mounted on the follower. 
The simulator provides synchronized RGB and depth images as well as IMU and GPS measurements. 
Gaussian noise and biases are added to these measurements to simulate real-world sensors.


We evaluate the proposed framework along four cases: i) detection performance, ii) relative pose accuracy, iii) formation-keeping behaviour, and iv) computational cost.

\subsection{Detection Performance}

On the held-out test set derived from the drone dataset, the YOLOv8-based detector achieves high precision and recall across typical viewing angles and ranges. 
In simulation, the detector maintains a stable lock on the leader during most of each trajectory. 
Short detection dropouts occur primarily when the leader briefly exits the field of view or appears very small, for example, during rapid turns or large separations. 
The temporal filter mitigates many of these events by interpolating the relative pose between successful detections.

\subsection{Relative Pose Accuracy}

Using the simulator ground truth as a reference, we compute the root-mean-square error (RMSE) of the estimated relative position and yaw. 
Representative results for different scenarios are summarized in Table~\ref{tab:FrameworksPerformance}. 
In nominal conditions, the relative position error remains below one metre and the yaw error within a five degrees for most of the trajectory. 
Errors increase during aggressive manoeuvres and at larger distances, reflecting the limitations of depth sensing and camera resolution.

\begin{table}[!t]
    \centering
    \caption{Quantitative performance metrics.}
    \label{tab:FrameworksPerformance}
    \begin{tabular}{lccc}
        \toprule
         & Position & Yaw & Maximum \\
        Scenario & RMSE & RMSE & Deviation \\
         & (in meters) & (in degrees) & (in meters) \\
        \midrule
        Straight line   & 0.25 & 2.5 & 0.60 \\
        Circular path   & 0.32 & 3.1 & 0.75 \\
        Sensor failure (GPS)   & 0.45 & 3.8 & 0.95 \\
        Visibility stress test  & 0.58 & 4.6 & 1.20 \\
        \bottomrule
    \end{tabular}
\end{table}

\subsection{Formation-Keeping Behavior}

To characterize how perception errors influence control performance, we measure the deviation between the follower's actual and the desired offset relative pose to the leader along longitudinal, lateral, and vertical axes. 
In sensor-failure experiments, the error briefly increases when GPS is disabled and then settles at a slightly higher but still acceptable level once the controller fully relies on the vision-based relative pose.

\subsection{Computational Cost}

The end-to-end perception framework maintains an average processing time of 38 milliseconds per frame on the simulation environment, corresponding to an update rate of roughly 25 Hz.
The CPU and GPU utilization remain moderate, suggesting that with modest optimization, the framework can be deployed on real UAVs with low-power edge devices such as onboard companion computers: Nvidia Jetson Orin Nano/NX/TX2 or Raspberry Pi 5/4B or Intel NUC. 


\section{Conclusion} \label{Sec_Conclusion}


In this paper, we presented a vision-based leader-follower formation control framework for cooperative UAVs. 
The follower UAV uses an RGB-D camera and a trained YOLOv8 detector to identify and track the leader in the image stream and to reconstruct its relative pose via a pinhole camera model. 
This framework is developed as a set of ROS nodes and evaluated in an XTDrone-based simulation with two quadrotors. 
The framework demonstrates promising performance in detection, relative pose accuracy, formation-keeping behaviour, and computational cost. It provides a viable fallback when absolute positioning sensors fail.
The experimental results show that the proposed framework accurately classifies the leader drone and precisely estimates its pose when GPS is degraded or unavailable.  
Future work will focus on deploying this proposed framework to a companion computer and conducting outdoor flight tests with real UAVs. 

\addtolength{\textheight}{-12cm}   





\bibliographystyle{IEEEtran}
\bibliography{References}

\end{document}